# Isoelastic Agents and Wealth Updates in Machine Learning Markets

Amos J. Storkey  A.STORKEY@ED.AC.UK
Jonathan J. Millin  J.J.MILLIN@SMS.ED.AC.UK
Krzysztof J. Geras  K.J.GERAS@SMS.ED.AC.UK
School of Informatics, University of Edinburgh

## Abstract

Recently, prediction markets have shown considerable promise for developing flexible mechanisms for machine learning. In this paper, agents with isoelastic utilities are considered. It is shown that the costs associated with homogeneous markets of agents with isoelastic utilities produce equilibrium prices corresponding to alpha-mixtures, with a particular form of mixing component relating to each agent's wealth. We also demonstrate that wealth accumulation for logarithmic and other isoelastic agents (through payoffs on prediction of training targets) can implement both Bayesian model updates and mixture weight updates by imposing different market payoff structures. An iterative algorithm is given for market equilibrium computation. We demonstrate that inhomogeneous markets of agents with isoelastic utilities outperform state of the art aggregate classifiers such as random forests, as well as single classifiers (neural networks, decision trees) on a number of machine learning benchmarks, and show that isoelastic combination methods are generally better than their logarithmic counterparts.

## 1. Introduction

This paper addresses the problem of classifier aggregation through the use of Machine Learning Markets. In supervised machine learning, many algorithms use simple averaging, weighted averaging, mixtures, mixing and log opinion pools for combining a variety of classifiers to form aggregate classifiers. On the other hand, prediction markets have been used for aggregation of simple beliefs in other fields (the market price is used as an aggregate probability). Some simple forms of theoretical prediction markets have been shown to be equivalent to weighted averaging (Barbu & Lay, 2011; Storkey, 2011). However, more generally, we can formulate prediction market mechanisms on the basis of expected utility theory. We show that a whole spectrum of aggregation methods are then available, including and extending those provided by $\alpha$-mixtures. We demonstrate the empirical benefits of these aggregation methods over those discussed above.

A number of authors have recently examined the possibilities of market mechanisms for implementing machine learning methods. Desirable properties of markets include the fact that agents act independently, markets are inherently parallelisable and the fact that the form of models produced by markets is more general than explicitly defined probabilistic models. Market based machine learning approaches are being shown to be consistent with standard probabilistic machine learning formalisms (Storkey, 2011) as well providing regret bounds for market makers (Chen & Wortman Vaughan, 2010; Abernethy et al., 2011). Markets are also a well studied and pervasive part of our social and computational infrastructure.

We envisage the possibility of setting up online prediction markets for large scale multivariate prediction problems in which different algorithms compete. This generalises prediction markets, which usually focus on disconnected discrete events, and relates to other recent work in utilising algorithm crowdsourcing for machine learning (Abernethy & Frongillo, 2011). It would provide an alternative to, say, the current interest in competition and challenge environments (e.g. Netflix, Kaggle, PASCAL2), where individuals compete for the best personal performance. Experience in these domains has suggested that it is common for individual algorithms to be outperformed by competitors grouping together to produce model combinations (Bell & Koren, 2007). This mirrors the experience regarding aggregation of expert predictions (Dani et al., 2006).

The main novel contributions of this paper are:

- Extending the set of standard agents from logarithmic agents and exponential agents to also include the various forms of isoelastic agents, and to show such agents can reproduce and extend the $\alpha$-mixtures framework (Amari, 2007).

- Demonstrating that online and batch agent wealth updates are equivalent to Bayesian posterior updates and mixture coefficient updates re-

spectively.

- Equating the multi-agent equilibrium (fixed-point) market with a divergence minimisation. The result is an iterative process for establishing the market equilibrium values.

- Demonstrating that Machine Learning Markets outperform individual strong classifiers when presented with equivalent data. Showing that the markets improve on state of the art classifiers, such as random forests, and provide better log-probability performance on classification test sets.

- Demonstrating that inhomogeneous sets of isoelastic agents produce better performance than logarithmic agents. Hence aggregation by simple mixtures can be improved upon.

## 2. Other Related Work

A number of previous papers address machine learning and prediction markets. In (Lay & Barbu, 2010) and (Barbu & Lay, 2011), the authors consider agents endowed with betting functions, and do experimental tests in the context of classifiers (leaves of the trees of a random forest) trained on bootstrapped samples. In (Storkey, 2011) the author develops a complementary approach, utilizing beliefs and utilities of individual agents. He shows that a variety of machine learning model combination methods, including model averaging, product models, factor graphs etc. can be implemented using Machine Learning Markets. In (Chen & Wortman Vaughan, 2010) the market is defined with respect to a global cost function, and they demonstrate achieving 'no regret' learning using information markets. From another perspective, Agrawal et al. (Agrawal et al., 2010) looked at the equilibrium conditions of prediction markets in various situations of matching buyers and sellers. The available information is important for market conditions. This was discussed in (Jumadinova & Dasgupta, 2011), where they used a multi-agent system to examine the various dependencies on information reliability, rate etc. It is the ability of prediction markets to aggregate belief that is key to their potential (Pennock & Wellman, 1997; Ottaviani & Sørensen, 2007), and much of the experimental work on prediction markets backs that up (Ledyard & Hanson, 2008).

The potential of prediction markets has been considered for some time. In (Arrow et al., 2008; Manski, 2006; Wolfers & Zitzewitz, 2004) the authors discuss the capability of practical prediction markets to capture accurate probabilities. In (Dani et al., 2006) the authors compared a number of different mechanisms for expert aggregation including a simple prediction market approach. Different market designs have different features, and ensuring good prediction market design with sufficient fluidity (Brahma et al., 2010) will be critical for efficiently reaching equilibrium. In (Tseng et al., 2010) the authors examine the statistical properties of market agent models, whereas in (Lee & Moretti, 2009) the authors consider prediction markets in the context of Bayesian learning.

## 3. Notation

Machine Learning Markets use prediction market mechanisms to generate machine learning models via the market price. The basic concepts and notation are now introduced.

### 3.1. Goods

We consider a set of *market goods* enumerated by $k = 1, 2, \ldots N_G$, each corresponding to a specific outcome of a discrete random variable, denoted $\mathsf{k}$. The good $k$ will pay out one unit of currency in the event that the outcome for $\mathsf{k}$ is $k$. The market has a commonly agreed cost $c_k$ ($0 < c_k < 1$) for each good $k$ and we collect the costs into the cost vector $\mathbf{c} = (c_1, c_2, \ldots, c_{N_G})^T$. The cost vector $\mathbf{c}$ will be interpreted as the aggregate probabilistic belief provided by the market: the probability[1] of $k$ occurring is $c_k$.

### 3.2. Agent Actions

A number of *agents*, enumerated by $i = 1, 2, \ldots, N_A$, act in the market. Each agent has wealth $W_i$, and invests (or *risks*) an amount $r_{ik}$ in stock $k$. We collect $\mathbf{r}_i = (r_{i1}, r_{i2}, \ldots, r_{iN_G})^T$. A no arbitrage assumption[2] implies that $\sum_k c_k = 1$, meaning a probabilistic interpretation of $c_k$ is reasonable. Likewise, we can require all agents to spend all their wealth: if any agent wants to keep it as a risk free investment, that agent can simply purchase one of each stock instead. Hence, without loss of generality, we have

$$\sum_k r_{ik} = W_i. \qquad (1)$$

Again without loss of generality, we will measure wealth in units such that the total wealth across all agents is 1: $\sum_i W_i = 1$. Hence $\sum_{ik} r_{ik} = 1$.

Each agent is also endowed with a utility function $U_i(W)$ denoting the utility of having wealth $W$. For the purposes of this paper we will only consider concave utility functions. Finally, each agent has a belief $P_i$, where $P_i(k)$ denotes the probabilistic belief, for that particular agent $i$, that the outcome of $\mathsf{k}$ will be $k$. Necessarily, as $k$ enumerates all possible outcomes, $\sum_k P_i(k) = 1$ for all agents.

---

[1] Strictly, the probability is $c_k$ divided by the unit of payout. This ensures dimensional consistency, both here and elsewhere.

[2] No arbitrage: there is no opportunity for a risk free gain. If $\sum_k c_k \neq 1$ an agent can make a risk free win by buying (or selling) one of each stock which has a sure return (or debt) of 1 unit.

### 3.3. Market

The agents jointly act in a market. The market transactions are subject to the macroscopic constraint

$$\sum_{i=1}^{N_A} \frac{r_{ik}}{c_k} = \sum_{i=1}^{N_A} W_i = 1 \Rightarrow \sum_{i=1}^{N_A} r_{ik} = c_k \quad (2)$$

where $N_A$ is the number of agents. This states that wealth must be conserved in the market: the total payout were that item to occur matches the total original wealth. Here $r_{ik}/c_k$ is the amount of good $k$ bought by agent $i$ (the amount invested divided by cost) and so is the amount received if event $k$ occurs.

### 3.4. Summary Table

| $i \to$ agent | $k \to$ good/outcome |
|---|---|
| $N_A, N_G \to$ #agents/goods | $\mathbf{r}_i \to$ investment |
| $W_i \to$ Wealth | $U_i \to$ Utility |
| $\mathbf{c} \to$ cost vector | $P_i(k) \to$ agent belief |

## 4. Utility Maximisation

In Machine Learning Markets the market price $\mathbf{c}$ defines a probability distribution over possible outcomes, which can be used for prediction. The multiclass prediction problem is the focus for this paper.

### 4.1. Investment Functions

A given utility function $U_i$ induces a given investment function $\mathbf{r}_i^*(W_i, \mathbf{c})$ via expected utility maximisation:

$$\mathbf{r}_i^* = \mathbf{r}_i^*(W_i, \mathbf{c}) = \arg\max_{\mathbf{r}_i} \sum_k P_i(k) U_i \left( \frac{r_{ik}}{c_k} \right)$$

$$\text{s.t.} \sum_k r_{ik} = W_i. \quad (3)$$

where we have used the fact that every agent must spend their whole wealth (1). The investment function indicates the amount an agent ideally would *wish* to invest in each good, given the costs $\mathbf{c}$. The market constraints may mean this desire cannot be satisfied: the agent must find a buyer or seller to realise this desire. The optimum of (3) is given by

$$r_{ik}^* = c_k (U_i')^{-1} \left( \lambda_i(\mathbf{c}) \frac{c_k}{P_i(k)} \right) \quad (4)$$

where $U'$ is the derivative of $U$ and $\lambda_i(\mathbf{c})$ is a Lagrange multiplier such that $\sum_k r_{ik}^* = W_i$ is satisfied.

In general, the equation for $\lambda_i$ cannot be explicitly solved. However, for a number of utilities the investment function is analytic. Table 1 lists some important utility functions (exponential, logarithmic and isoelastic) and their corresponding investment functions.

The class of isoelastic functions are a very useful set of utilities. The isoelastic utilities get their name as they all have investment functions that are linear in the current wealth; this property is called the *isoelastic property*. Isoelastic utility functions are parameterised by $\eta > 0$. Strictly, the logarithmic utility is also an isoelastic utility with the limiting value of $\eta = 1$.

### 4.2. Market Equilibrium

The Market must satisfy the market constraint (2). At the same time each agent attempts to maximise their individual utility. It is well known (Arrow & Debreu, 1954) that, if the individual utilities are concave (as is the case in this paper), there is a unique fixed price point for which agents all attain their maximum utility and the market constraints are satisfied. This is called the *market equilibrium*. However, the existence of a fixed point does not establish a means of obtaining it. The question of how a market might equilibrate formed part of the early discussion regarding equilibria, and led to Walras' concept of *tâtonnement*. This idea, as communicated by Samuelson (Samuelson, 1947), was that prices are differentially changed in the direction of the excess or deficit demand. However, the constraints on this formalism meant it was not established as a general purpose procedure. The are many analyses that involve formulating convex optimisation approaches or auction processes for obtaining market equilibria, e.g. (Deng et al., 2002; Devanur et al., 2008; Ye, 2006) – see (Vazirani, 2007) for more details. These algorithms typically require a complete optimisation procedure. Two exceptions are (Cole & Fleischer, 2007; Fleischer et al., 2008) which develop on the idea of tâtonnement.

In this paper an iterative tâtonnement-like approach is used for establishing market equilibria. Consider the fact that $\sum_k c_k = 1$ and $\sum_k (\sum_i r_{ik}) = 1$ means that both $c_k$ and $\sum_i r_{ik}$ take probabilistic form. At equilibrium, we have $\sum_i r_{ik} = c_k$ (see (2)) when all agents are allocated their optimal demand. Away from equilibrium there will be an excess or deficit demand in different goods, which is evident in the difference between $\sum_i r_{ik}$ and $c_k$. Consider the KL divergence $KL(\mathbf{c} || \sum_i \mathbf{r}_i)$. This is minimised and zero only at equilibrium. To minimise this KL divergence we use Algorithm 1. This algorithm only terminates when the (unique) equilibrium is reached, when $\mathbf{c}$ is the equilibrium price. In all our empirical tests the equilibrium was always reached. Each pass of the algorithm is naively $O(N_A \times N_G)$ – each update is computationally equivalent to a mixture model update. In our experiments equilibria were reached in between 5 and 15 iterations.

Satisfaction of the market constraints for given buying functions, or equivalently minimisation of $KL(\mathbf{c} || \sum_i \mathbf{r}_i)$, defines a fixed point that is a market equilibrium. These market equilibria can be explicitly computed for various agent utilities. For a market of identical logarithmic agents or a market of identical exponential agents we have

$$c_k = \frac{\sum_i W_i P_i(k)}{\sum_i W_i} \text{ and } c_k \propto \prod_{i=1}^{N_A} P_i(k)^{\frac{1}{N_A}}, \quad (6)$$

| | | |
|---|---|---|
| **Exponential** | $U(W) = -\exp(-W)$ | $r_{ik}(W_i, \mathbf{c}) = \frac{W_i}{N_G} + c_k \log \frac{P_i(k)}{c_k} - \frac{1}{N_G}\sum_{k'} c_{k'} \log \frac{P_i(k')}{c_{k'}}$ |
| **Logarithmic** | $U(W) = \begin{cases} \log(W) & \text{for } W > 0 \\ -\infty & \text{otherwise} \end{cases}$ | $r_{ik}(W_i, \mathbf{c}) = W_i P_i(k)$ |
| **Isoelastic** | $U(W) = \frac{W^{1-\eta_i} - 1}{1-\eta_i}$ | $r_{ik} = W_i \left( \dfrac{(c_k)^{\frac{\eta_i - 1}{\eta_i}} (P_i(k))^{\frac{1}{\eta_i}}}{\sum_{k'} (c_{k'})^{\frac{\eta_i - 1}{\eta_i}} (P_i(k'))^{\frac{1}{\eta_i}}} \right)$ |

Table 1. Various utility functions and their corresponding investment functions. For the isoelastic utilities $0 < \eta$.

**Algorithm 1** Market Equilibrium

initialise $\mathbf{c}$, initialise $a$ (e.g. $a = 0.1$)
define stopping criterion $\epsilon$
**repeat**
  Compute optimal $\mathbf{r}_i$ for each agent ignoring market constraint
  set (for normalising $Z_1$)
$$c_k^{\text{new}} = \frac{1}{Z_1} \left( \frac{\sum_i r_{ik}}{c_k} \right)^{1-a} c_k \quad (5)$$
  **if** $\text{KL}(\mathbf{c}^{\text{new}} \| \sum_i \mathbf{r}_i(\mathbf{c}^{\text{new}})) - \text{KL}(\mathbf{c} \| \sum_i \mathbf{r}_i(\mathbf{c})) < 0$ **then**
    discard $c_k^{\text{new}}$ and increase $a$.
  **end if**
**until** $\text{KL}(\mathbf{c}^{\text{new}} \| \sum_i \mathbf{r}_i(\mathbf{c}^{\text{new}})) < \epsilon$

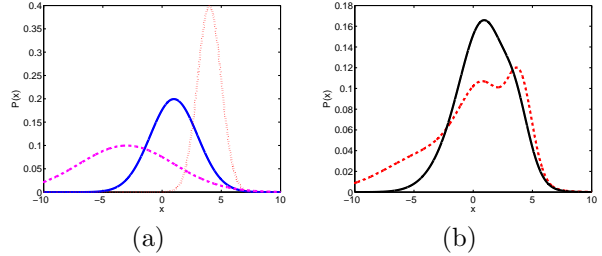

Figure 1. (a) Three different components (i.e. agent beliefs), each given weights $W_i$ of 0.4, 0.4 and 0.2 from left to right. (b) The logarithmic (i.e. mixture) combination of these components (dashed) and the isoelastic ($\eta = 10$) combination (solid). Note the isoelastic combination puts more weight where the overlap of the different components are and down-weights the regions of disagreement or isolated components.

respectively. For logarithmic agents this takes the form of a model average or mixture of the agents' beliefs; for exponential agents it is a log opinion pool of beliefs. See (Storkey, 2011) for discussion of these. The term *homogeneous* will be used to refer to markets where all agents have identical utility functions.

### 4.3. Equilibria for Isoelastic Agents

We cannot explicitly obtain the equilibrium for sets of isoelastic agents with $\eta \neq 1$, and so the optimisation procedure of the previous section needs to be employed. However we can say something about the form of solution we obtain for isoelastic agents. For example we can obtain the following equilibrium for homogeneous markets of isoelastic agents, all having the same $\eta$. This solution is not closed form:

$$c_k = \left[ \sum_i V_i P_i(k)^{\frac{1}{\eta}} \right]^{\eta} \quad (7)$$

where $V_i = W_i/Z_i$ and $Z_i$ is the implicit solution to

$$Z_i = \sum_k \left[ \sum_j \frac{W_j}{Z_j} P_j(k)^{\frac{1}{\eta}} \right]^{\eta - 1} P_i(k)^{\frac{1}{\eta}}. \quad (8)$$

Equation (7) is precisely the equation for $\alpha$-mixtures (Amari, 2007; Wu, 2009), but where $V_i$ is defined implicitly in terms of a set of weights (or wealths) $W_i$. This particular expression for $\alpha$-mixtures is interesting because of the isoelastic property: if a single component from the $\alpha$-mixture is replaced by two identical components with weight $W_i/2$, it results in exactly the same model. The precise number of components/agents does not matter, just the total wealth associated with each belief, regardless of the how many components share it. Furthermore the influence the agent has on the overall model depends on the level of agreement that the agent has with the overall consensus. This affect the value of $Z_i$ for that agent.

Figure 1 illustrates the distinction between an isoelastic market combination and a logarithmic combination (which is equivalent to a standard mixture). In the isoelastic market, for $\eta > 1$, the individual beliefs are 'squashed' (raised to a fractional power) before being mixed, and are then 'unsquashed' again after mixing. The result of this is the areas of agreement between agents are emphasised relative to a standard mixture.

As an alternative to (7), we can also write

$$c_k = \sum_i W_i P_{ik}^{\eta}(\mathbf{c}) \quad (9)$$

where $P_{ik}^{\eta}(\mathbf{c})$ is defined as

$$P_{ik}^{\eta}(\mathbf{c}) = \frac{c_k \left( \frac{P_i(k)}{c_k} \right)^{1/\eta}}{\sum_{k'} c_{k'} \left( \frac{P_i(k')}{c_{k'}} \right)^{1/\eta}}. \quad (10)$$

Again this is not closed form, but expresses the equilibrium $c_k$ as a weighted sum of the *effective beliefs* $P_{ik}^{\eta}$ that are associated with each agent once the impact of the combination with rest of the market is taken into account. Each effective belief is weighted by the agent's wealth $W_i$ before aggregation.

In the discussion above, all the agents have utilities with the same value of $\eta$. This is a homogeneous market structure. However there is no restriction to homogeneity in the context of Machine Learning Markets. Here an *inhomogeneous* market structure can be used where different agents have different $\eta$ values. Hence the equilibria of Machine Learning Markets can implement a broader set of combination processes than standard $\alpha$-mixtures, and so generalises the $\alpha$-mixture formalism. We show in Section 6 that the use of an inhomogeneous market structure provides improvements over standard mixtures for a variety of standard machine learning benchmarks.

## 5. Training and Wealth Allocation

We consider a classification problem, with a training dataset $D_{\text{Tr}}$ consisting of covariates $\mathbf{x}^n$ and corresponding classifications $k^n$. We wish to learn the relationship between variables $\mathbf{x}$ and classifications $k$, so that given a test point $\mathbf{x}^*$ we can provide a predictive distribution for the value that the corresponding class label $k^*$ will take. A test dataset $D_{\text{Te}}$ is used to evaluate the final performance.

### 5.1. Agent Beliefs and Wealth Updates

Each agent has an individual belief $P_i(k)$. In this paper we consider agents that have used standard machine learning algorithms on a training data set $D_{\text{Tr}}$ to derive their beliefs.

Each agent has a specific wealth $W_i$. The wealth affects the market influence of that agent. An agent's action in the market changes the wealth of the agent: the change of wealth after a single investment $\mathbf{r}_i$, and a return on that purchase is given by $r_{ik^*}/c_{k^*}$ where $k^*$ denotes the index of the event that occurs. The agents make purchases in predictions across the whole training dataset, trading with the other agents. Payouts are then made and each agent makes a return on the investment. Agents that invest well (relatively) gain wealth, whereas agents that invest in goods that don't pay out lose wealth and hence market influence. This process can be repeated for a number of *epochs*.

For analytic purposes, we consider two wealth update schemes: an online and a batch scheme. All the empirical analyses are done using the batch scheme.

#### 5.1.1. ONLINE: BAYESIAN MODEL UPDATES

Consider the case of all agents starting with wealth $1/N_A$, where $N_A$ is the number of agents. Let $T$ denote the first $t$ data items and $D_T$ denote the ordered set of those data items, with $D_t$ being the $t$th item. The total number of items is $N_{\text{Tr}}$. Let $k_t$ be the target for the $t$th data point.

In the online setting, the agent purchases predictions on the data points $t = 1, 2, \ldots, N_{\text{Tr}}$ one at a time. At each time point the outcome is then revealed and all bets are cashed in. This is an *online* update scheme. Let $W_i^t$ denote the wealth of agent $i$ after the target for data point $t$ is known and the winnings are received.

For isoelastic agents (including logarithmic agents when $\eta = 1$), at each data point $t$, each agent bets the whole wealth $W_i^t$ and gains a return of $r_{ik_t}/c_{k_t}$, leading to

$$W_i^{t+1} = \frac{W_i^t P_{ik_t}^\eta(\mathbf{c})}{\sum_{i'} W_{i'}^t P_{i'k_t}^\eta(\mathbf{c})} \quad (11)$$

using (10) and (9), and $r_{ik}$ from Table 1.

If we equate $W_i^{t+1}$ with the concept $P(i|D_{t+1})$ then this leads to

$$W_i^{t+1} = P(i|D_{T+1}) = \frac{P_{ik_t}^\eta P(i|D_T)}{\sum_{i'} P_{i'k_t}^\eta P(i'|D_T)} \quad (12)$$

which gives the Bayesian update rule on observation of a new data point at time $t$ for a single agent likelihood $P_{ik_t}^\eta$. For logarithmic agents the belief $P_{ik_t}^\eta = P_i(k_t)$ and this just reduces to a Bayes update, treating each agent as an independent probabilistic model. For isoelastic agents it is still a Bayes update, but where Bayes rule uses the *effective beliefs* $P_{ik_t}^\eta$ as the component distributions, instead of the individual agent belief $P_i(k_t)$. The equilibrium cost $\mathbf{c}^t$ for any item at time $t$ after seeing data $D_T$ is a standard Bayesian model average, given by (9), as the weights are the posterior probabilities associated with each agent (12).

The fact that these Bayesian updates occur for logarithmic utilities (or equivalently log-loss) has been discussed in a different context in (Beygelzimer et al., 2012). Establishing the extension of this rule for isoelastic agents is a novel generalisation.

#### 5.1.2. BATCH: MIXING COEFFICIENT UPDATES

Bayesian model averaging is appropriate if we interpret each agent as an alternative competing hypothesis, where ultimately one agent has the correct belief. However, in many, or even most situations (see e.g. (Domingos, 2000; 1997; Minka, 2002) for a continued discussion), we may believe that the most appropriate model is a combination of beliefs rather than a single one. In those settings Bayesian model averaging is inappropriate. Rather, we may believe the data is best described by a mixture of probabilities, and we wish to determine optimal mixing proportions.

Consider, instead, splitting the agent wealth equally across test cases, and requiring the agents to place bets on all test cases at once. In this case the wealth updates are equivalent to a single step of the mixture component updates. Specifically the return from considering data item $t$ is

$$P(i|t) \stackrel{\text{def}}{=} \frac{W_i P_{ik_t}^\eta(\mathbf{c})}{\sum_{i'} W_{i'} P_{i'k}^\eta(\mathbf{c})}. \quad (13)$$

Equation (13) is precisely the form of responsibility

Isoelastic Agents and Wealth Updates in Machine Learning Markets

|  | Tests | Waveform 21f, 3c, 5000d | Vehicle 18f, 3c, 946d | Image 19f, 7c, 2310d | Ionosphere 34f, 2c, 351d | Breast Cancer 10f, 2c, 699d | Sonar 60f, 2c, 208d | Letter Recognition 16f, 26c, 20000d |
|---|---|---|---|---|---|---|---|---|
| RF | LLR | 30 | 7.9 | 21 | 1.9 | 2 | 1.1 | 190 |
|  | $\sigma$(LLR) | 5.9 | 2.4 | 3.1 | 1.9 | 1.9 | 0.8 | 11 |
|  | p | $6.8 \times 10^{-23}$ | $1.1 \times 10^{-17}$ | $1.9 \times 10^{-26}$ | $3.4 \times 10^{-6}$ | $1.2 \times 10^{-6}$ | $6.4 \times 10^{-9}$ | $3.9 \times 10^{-38}$ |
| isoNN | LLR | −41 | −31 | −8 | 5.6 | 0.5 | 2.2 | 210 |
|  | $\sigma$(LLR) | 15 | 9.8 | 7.9 | 8 | 3.8 | 5.1 | 29 |
|  | p | 1 | 1 | 1 | 0.00031 | 0.22 | 0.012 | $3.4 \times 10^{-27}$ |
| NN | LLR | 28 | 12 | 27 | 19 | 20 | 16 | 390 |
|  | $\sigma$(LLR) | 33 | 21 | 19 | 11 | 10 | 7.9 | 41 |
|  | p | $3.7 \times 10^{-5}$ | 0.0024 | $5.1 \times 10^{-9}$ | $5.4 \times 10^{-11}$ | $4.8 \times 10^{-12}$ | $6.2 \times 10^{-12}$ | $1.5 \times 10^{-30}$ |
| isoDT | LLR | 28 | 7.3 | 16 | 2.1 | 3.7 | 1.4 | 12 |
|  | $\sigma$(LLR) | 13 | 5.8 | 8.8 | 3.2 | 3.4 | 2.6 | 28 |
|  | p | $5.2 \times 10^{-13}$ | $6.9 \times 10^{-8}$ | $2.4 \times 10^{-11}$ | 0.00056 | $8.8 \times 10^{-7}$ | 0.0035 | 0.01 |
| DT | LLR | $2.3 \times 10^2$ | 42 | 61 | 6.8 | 14 | 8.8 | 580 |
|  | $\sigma$(LLR) | 30 | 13 | 18 | 4.2 | 7.2 | 3.6 | 73 |
|  | p | $6.5 \times 10^{-28}$ | $2.7 \times 10^{-17}$ | $3.9 \times 10^{-18}$ | $4.4 \times 10^{-10}$ | $1.4 \times 10^{-11}$ | $2.8 \times 10^{-14}$ | $3.1 \times 10^{-28}$ |

*Table 2.* **Comparisons against isoRF**. The table uses an isoelastic market of trees from a random forest (isoRF) as a baseline and compares this with a number of methods above (positive LLR => isoRF is better). The isoelastic market of trees (isoRF) performs better than the other standalone classifiers and random forest aggregation which is among state of the art on all these problems (Caruana & Niculescu-Mizil, 2006). The exception is that isoRF is beaten in some cases by the isoelastic market of neural networks. The other methods listed are the standalone random forest (RF), an inhomogeneous isoelastic market of neural networks (isoNN), a standalone neural network (NN), an inhomogeneous isoelastic market of decision trees (isoDT) and a standalone decision tree (DT). LLR is the average test log likelihood-ratio (difference of log likelihoods) between the isoRF and other listed methods. $\sigma$(LLR) is the standard error of that log likelihood-ratio across different data samples. The $p$ values give the sampling probability of each method being better than isoRF using a right-tailed paired t-test with 29 degrees of freedom. The information about the dataset is displayed below its name, where $f$ is the number of features, $c$ the number of classes, and $d$ the number of data points. Comparing test log likelihood takes into account the full prediction probabilities. It captures not just the highest probability class but the quality of the measure of uncertainty across classes.

calculation for a mixture model, but where we have used the effective beliefs for the isoelastic agents. The update rule $W_i = \sum_t P(i|t)$, which is simply the accumulated return over the whole dataset, matches the update rule for mixture coefficients.

Hence across the whole range of isoelastic agents, initialising agents with equal wealth and repeatedly applying the batch update rule reproduces the usual mixture coefficient updates applied to an $\alpha$-mixture model. However multiple different values of $\alpha$ can be used for different agents.

## 6. Results

Machine Learning Markets with logarithmic and isoelastic agents were compared with decision trees, neural networks and random forests on a number of UCI datasets[3]. Experimental data was split into two sets, with 2/3 of the data being used for training and 1/3 for testing, with a maximum total dataset size of 3200 items. For large multiclass data, we used the *Letter Recognition* dataset. The markets of random forests were built using the MATLAB random forest implementation, `treebagger`, with 20 decision trees pruned by requiring a minimum of ten of observations per tree leaf. Individual decision trees were extracted from the random forest after it has been trained on all of the training data, and were used to generate each of the 20 agents' beliefs. In our comparisons, market wealths were adapted on the complete training set with 1 training epoch. 30 iterations of each test were performed to generate meaningful statistics, with data being randomly permuted before each test. The same series of random seeds were used for each iteration of each test in order to fairly compare different utility functions. Wealth updates were performed using the batch mechanism described in Section 5.1.2.

Inhomogeneous isoelastic markets were created by sampling values for $\eta$ using $(\eta - 1) \sim \Gamma(k, \theta)$, with shape parameter $k = 3$ and scale parameter $\theta = 1$. This produces a diverse set of $\eta$ values for different agents, while ensuring $\eta > 1$. More risk averse utility functions ($\eta > 1$) were chosen as they emphasise regions of agreement between agents rather than regions of disagreement (see (7) and Figure 1).

The primary purpose of this analysis is to test different possible probabilistic combination methods against other single classifiers. Hence, we compute test log-likelihoods as the main evaluation metric. This is given by $LL = \sum_t^T \log(P(k_t|\mathbf{x}_t))$, where $P(k_t|\mathbf{x}_t)$ is the probability of the true value $k_t$ given covariates $\mathbf{x}_t$. For a market we have $P(k_t|\mathbf{x}_t) = c_{k_t}$ where $c$ is the equilibrium cost from the market given all agents in the market know the covariates $\mathbf{x}$. Logs of test likelihood ratios are used when different models are compared.

### 6.1. Relative Classifier Performance

We compare the methods used in Machine Learning Markets against other standard classifiers. Machine Learning Markets can utilise any probabilistic classifier as the beliefs for each agent, and so we compare a number of single classifiers with a market of those clas-sifiers. The exception is for random forests, which is already an aggregate classifier, and there we compare a random forest, against a market of trees that match the trees in the random forest. An inhomogeneous isoelastic market with batch updates is used for all comparisons. Table 2 presents the results along with standard deviations. The market approaches outperform all the standard approaches on all datasets, with clear statistical significance. This includes improvements over random forests. Interestingly, the market of neural networks performed better than the market of random trees in some settings.

---
[3]Available at: http://archive.ics.uci.edu/ml/

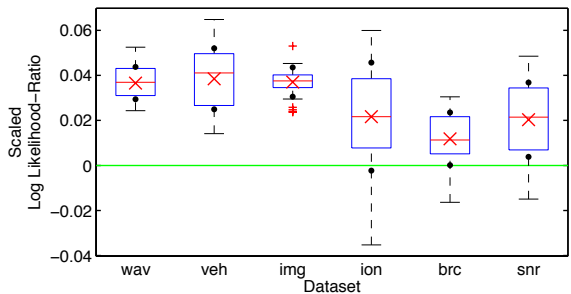

*Figure 2.* A boxplot (the red 'x' represents the mean, and black dots represent one standard deviation from the mean) of the scaled test log likelihood-ratios in the isoelastic and logarithmic markets for many iterations, scaled by the number of data points. Note that in general, the log-likelihood for isoelastic agents is significantly higher than for logarithmic agents. Letter recognition is excluded as it would be off the top of the figure.

### 6.2. Isoelastic versus Logarithmic Markets

Figure 2 shows $\frac{1}{N_{Te}} \log \left( \frac{L^{\text{ISO}}}{L^{\text{LOG}}} \right)$, where $N_{Te}$ is the number of test points, and $L$ denotes the test likelihood. This is referred to as the (scaled) log of the test likelihood-ratio between the inhomogeneous isoelastic market and the logarithmic market predictions.

In general, the log likelihood-ratios are positive, meaning that isoelastic markets have higher test performance than logarithmic markets. Further, they are positive to one standard deviation (the black dots in Figure 2), implying that isoelastic markets perform better (paired t-test $p < 0.01$ in all cases).

### 6.3. Varying the Parameter of Isoelasticity

Figure 3 shows that the test log-likelihood varies for homogeneous isoelastic markets with varying $\eta$. Searching for a good $\eta$ via a cross validation process can be computationally expensive. An alternative approach is to randomly allocate an $\eta$ to each of the agents, producing inhomogeneous markets, and perform market updates in order to tune the mixing proportions for the different agents. Figure 3 demonstrates that the inhomogeneous market provides results about as good as if we has known the optimal test $\eta$ but with significantly less computational cost.

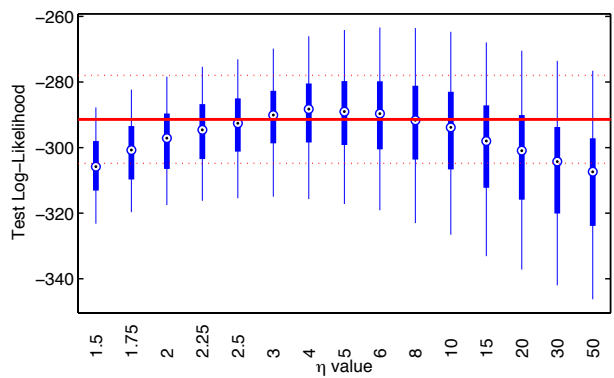

*Figure 3.* Boxplot showing the test log-likelihoods for different values of $\eta$ (on waveform dataset). The red line gives the test log-likelihood for the same data and an inhomogeneous market with $(\eta - 1) \sim \Gamma(3, 1)$. The inhomogeneous market performs on par with the market with the best $\eta$ value ('best' as assessed a posteriori on the test data), without prior knowledge of good $\eta$ values.

### 6.4. Batch Wealth Updates and Performance

Figure 4 shows that adapting wealth improves the test log-likelihood. This is true for both logarithmic and isoelastic utility functions. We have noticed that wealth adaptation does not make a significant difference on accuracy for small multiclass datasets, however, improvements in accuracy are observed on the large multiclass Letter Recognition data. In general, learning is more beneficial in cases where some agents are significantly poorer performers than others (e.g. they overfit, or are trained on biased data etc.).

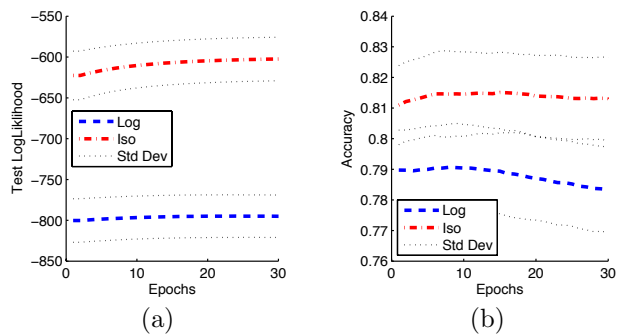

*Figure 4.* The (a) log likelihoods and (b) accuracy for isoelastic and logarithmic utility functions, over a number of training epochs for the Letter datasets. Value of log utility at time zero is equivalent to the value for a random forest. The isoelastic utility is better than logarithmic utility and random forest in both accuracy and log likelihood. The gain from using an isoelastic utility is greater than the gain from wealth adaptation (equivalent to mixture weight optimisation). The wealth adaptation does provide some benefit, but is more much more useful when spurious or poor classifiers are also included. Here most individual classifiers provide a similar contribution.

# 7. Discussion

Machine Learning Markets can reflect many of the properties of principled probabilistic methods in hand-crafted probabilistic models. The design of Machine Learning Markets allows the implicit definition of powerful models. We show that Baysesian model averaging and mixture model learning can be naturally implemented using market mechanisms. We show that different utility functions have a significant effect on the market combination results, and that isoelastic utilities are more effective in a number of tests than utilities that implement standard mixtures. The benefits of inhomogeneous markets of isoelastic agents over state of the art classifiers has been demonstrated, and the understanding of isoelastic utilities as encoding a generalisation of $\alpha$-mixtures has been developed.

There are two immediate extensions to this work to be considered. First, the adaptability of markets means that tests of this approach in the context of dataset shift, or non-stationary environments, would be valuable. Another angle worthy of investigation is the mixture of expert setting. In the context of this paper, all agents had beliefs about the whole predictive dataset. It is likely that an agent will also learn about its own performance in the market: assessing what situations it is likely to generate a positive return on. Such agents would allocate different resources to different conditional situations akin to a mixture of experts.

# References

Abernethy, J. and Frongillo, R. A collaborative mechanism for crowdsourcing prediction problems. In *Advances in NIPS 24 (NIPS2011)*, 2011.

Abernethy, J., Chen, Y., and Wortman Vaughan, J. An optimization-based framework for automated market-making. In *12th ACM Conference on Electronic Commerce (EC 2011)*, 2011.

Agrawal, S., Megiddo, N., and Armbruster, B. Equilibrium in prediction markets with buyers and sellers. *Economics Letters*, 109:46–49, 2010.

Amari, S. Integration of stochastic models by minimizing $\alpha$-divergence. *Neural Computation*, 19(10):2780–2796, 2007.

Arrow, K.J. and Debreu, G. Existence of an equilibrium for a competitive economy. *Econometrica: Journal of the Econometric Society*, 22(3):265–290, 1954.

Arrow, K.J. et al. The promise of prediction markets. *Science*, 320:877, 2008.

Barbu, A. and Lay, N. An introduction to artificial prediction markets for classification. arXiv:1102.1465v3, 2011.

Bell, R.M. and Koren, Y. Lessons from the Netflix prize challenge. *ACM SIGKDD Explorations*, 9(2):75–79, 2007.

Beygelzimer, A., Langford, J., and Pennock, D. Learning performance of prediction markets with kelly bettors. In *Proceedings of AAMAS 2012*, 2012.

Brahma, A., Das, S., and Magdon-Ismail, M. Comparing prediction market structures, with an application to market making. arXiv:1009.1446, 2010.

Caruana, R. and Niculescu-Mizil, A. An empirical comparison of supervised learning algorithms. In *Proceedings of ICML 2006*, 2006.

Chen, Y. and Wortman Vaughan, J. A new understanding of prediction markets via no-regret learning. In *Proceedings of the 11th ACM conference on Electronic commerce*, 2010.

Cole, R. and Fleischer, L. Fast-converging tatonnement algorithms for the market problem. Technical report, Dept. Computer Science. Dartmouth College., 2007.

Dani, V. et al. An empirical comparison of algorithms for aggregating expert predictions. In *Proceedings of the Conference on Uncertainty in AI (UAI)*, 2006.

Deng, X., Papadimitriou, C., Saberi, A., and Vazirani, V. On the complexity of equilibria. In *STOC 2002*, 2002.

Devanur, N.R., Papadimitriou, C.H., Saberi, A., and Vaziani, V.V. Market equilibrium via a primal–dual algorithm for a convex program. *Journal of the ACM*, 55:22:1–22:18, 2008.

Domingos, Pedro. Why does bagging work? a Bayesian account and its implications. In *Proceedings KDD*, 1997.

Domingos, Pedro. Bayesian averaging of classifiers and the overfitting problem. In *In Proc. ICML 2000*, 2000.

Fleischer, L., Garg, R., Kapoor, S., Khandekar, R., and Saberi, A. A fast and simple algorithm for computing market equilibria. In *Proceedings of WINE*, pp. 19–30, 2008.

Jumadinova, J. and Dasgupta, P. A multi-agent system for analyzing the effect of information on prediction markets. *International Journal of Intelligent Systems*, 26:383–409, 2011.

Lay, N. and Barbu, A. Supervised aggregation of classifiers using artificial prediction markets. In *Proceedings of ICML*, 2010.

Ledyard, J and Hanson, R. An experimental test of combinatorial information markets. *Journal of Economic Behavior and Organization*, pp. 469–483, 2008.

Lee, D.S. and Moretti, E. Bayesian learning and the pricing of new information: Evidence from prediction markets. *American Economic Review*, 99(2):330–336, 2009.

Manski, C.F. Interpreting the predictions of prediction markets. *Economics Letters*, 91:425429, 2006.

Minka, T. Bayesian model averaging is not model combination. Technical report, MIT Media Lab Note, 2002.

Ottaviani, M. and Sørensen, P.N. Aggregation of information and beliefs in prediction markets. FRU Working Papers, 2007.

Pennock, D.M. and Wellman, M.P. Representing aggregate belief through the competitive equilibrium of a securities market. In *Proceedings of the Thirteenth Conference on Uncertainty in AI*, pp. 392–400, 1997.

Samuelson, P.A. *Foundations of Economic Analysis*. Harvard University Press, 1947.

Storkey, A.J. Machine Learning Markets. In *Proceedings of Artificial Intelligence and Statistics*, volume 15. Journal of Machine Learning Research W&CP, 2011.

Tseng, J.J., Lin, C.H., Lin, C.T., Wang, S.C., and Li, S.P. Statistical properties of agent-based models in markets with continuous double auction mechanism. *Physica A: Statistical Mechanics and its Applications*, 389(8):1699–1707, 2010.

Vazirani, V. Combinatorial algorithms for market equilibria. In et al., N. Nisan (ed.), *Algorithmic Game Theory*. Cambridge University Press, 2007.

Wolfers, J. and Zitzewitz, E. Prediction markets. *Journal of Economic Perspectives*, 1:107–126, 2004.

Wu, D. Parameter estimation for $\alpha$-GMM based on maximum likelihood criterion. *Neural computation*, 21(6):1776–1795, 2009.

Ye, Y. A path to the Arrow-Debreu competitive market equilibrium. *Mathematical Programming*, 2006.